\documentclass[review]{elsarticle}

\usepackage{hyperref}

\journal{Journal of \LaTeX\ Templates}

\usepackage{amsmath,amssymb,amsfonts}
\usepackage{algorithm,algpseudocode}
\usepackage{graphicx}
\usepackage{textcomp}
\usepackage{xcolor}

\usepackage{amsthm}
\usepackage{booktabs}
\usepackage{subcaption}
\usepackage{multirow}
\usepackage{makecell}

\newcommand{\clh}{\color{black}}

\newcommand{\clhrev}{\color{black}}

\algrenewcommand\algorithmicrequire{\textbf{Input:}}
\algrenewcommand\algorithmicensure{\textbf{Output:}}

\begin{document}

\begin{frontmatter}

\title{Rule Mining over Knowledge Graphs via Reinforcement Learning}

\author[fudan]{Lihan Chen}
\ead[email]{lhc825@gmail.com}

\author[fudan]{Sihang Jiang}
\ead[email]{tedsihangjiang@gmail.com}

\author[fudan]{Jingping Liu}
\ead[email]{jpliu17@fudan.edu.cn}

\author[fudan]{Chao Wang}
\ead[email]{17110240038@fudan.edu.cn}

\author[ncsu]{Sheng Zhang}
\ead[email]{szhang37@ncsu.edu}

\author[fudan]{Chenhao Xie}
\ead[email]{redreamality@gmail.com}

\author[fudan]{Jiaqing Liang}
\ead[email]{l.j.q.light@gmail.com}

\author[fudan]{Yanghua Xiao\corref{mycorrespondingauthor}}
\cortext[mycorrespondingauthor]{Corresponding author}
\ead[email]{shawyh@fudan.edu.cn}

\author[ncsu]{Rui Song}
\ead[email]{rsong@ncsu.edu}

\address[fudan]{Fudan University, Shanghai, China}
\address[ncsu]{North Carolina State University, Raleigh, US}

\begin{abstract}
	Knowledge graphs (KGs) are an important source repository for a wide range of applications and rule mining from KGs recently attracts wide research interest in the KG-related research community. Many solutions have been proposed for the rule mining from large-scale KGs, which however are limited in the inefficiency of rule generation and ineffectiveness of rule evaluation. To solve these problems, in this paper we propose a generation-then-evaluation rule mining approach guided by reinforcement learning. Specifically, a two-phased framework is designed. The first phase aims to train a reinforcement learning agent for rule generation from KGs, and the second is to utilize the value function of the agent to guide the step-by-step rule generation. 
	We conduct extensive experiments on several datasets and the results prove that our rule mining solution achieves state-of-the-art performance in terms of efficiency and effectiveness.
\end{abstract}


\end{frontmatter}

\section{Introduction}
Knowledge graphs (KGs) such as FreeBase~\cite{bollacker2008freebase}, DBpedia~\cite{auer2007dbpedia}, and Wikidata~\cite{vrandevcic2014wikidata} draw much attention due to their wide range of applications including reasoning~\cite{wang2020adrl} and recommendation~\cite{tao2021multi}, etc. 
The usefulness of KGs lies in their large-scale high-quality well-structured facts which are usually extracted from text corpora.
Being large-scale and abundant, KGs themselves have become important source repositories from which we can mine high-quality symbolic knowledge, especially in the form of \textit{rules}.
For example, a rule $bornIn(x, y) \wedge country(y, z) \rightarrow nationality(x, z)$ can be mined from many existing KGs. This rule means that if person $x$ was born in city $y$ and $y$ is in country $z$, then $x$ is a citizen of $z$, where the left part is the body of the rule and the right is the head. \emph{Rule mining} or \emph{rule discovery} has become one of the most important ways to exploit the values of KGs~\cite{galarraga2015fast,barati2017mining,chiclana2018arm}.
In this paper, we focus on the \emph{efficient} and \emph{effective} rule mining from KGs.

\subsection{Background}
Rule mining from KGs recently attracts wide research interest in the KG-related research community due to the following reasons.
First, rules provide an explainable way to derive new facts to complete an existing KG. Compared to deep learning models, rules are explicit symbolic knowledge that are human-understandable.
Second, rules mined from KG are also helpful for identifying potential errors in the KG. For example, a fact contradicting a high-confidence rule is very likely to be wrong.
Third, rules improve the reasoning capability of a KG. Many reasoning approaches rely on rules provided by other upstream methods~\cite{galarraga2015fast}. 
Last but not the least, rules characterize the regularity of the data, thus providing more opportunities for efficient KG data management.


Many solutions have been proposed for rule mining from large-scale KGs, which however have great room to be improved in terms of \emph{efficiency} and \emph{effectiveness}. These solutions usually take the \emph{generation-then-evaluation} framework~\cite{galarraga2015fast,omran2018scalable}.
For example, given a head predicate (say $nationality(x,y)$), traditional rule miners first generate all possible rules within a certain length with the head predicate and then evaluate their quality to find good rules (such as $bornIn(x,z) \land country(z,y)\rightarrow nationality(x,y)$).
These solutions have several weaknesses.
First, the brutal force enumeration is obviously \emph{inefficient} and cannot scale up to a large KG that contains millions of facts and thousands of predicates.
Second, previous measures for evaluating rules are still weak in \emph{effectiveness}. 
For example, the most widely used statistical measures are \emph{support} and \emph{confidence}~\cite{galarraga2015fast}.
These measures evaluate rules by directly counting the number of instances existing in KGs.
Their effectiveness is undermined by the incompleteness of KGs. 
These statistic measures tend to bias towards popular facts while underestimating rarely observed facts. 
A lot of follow-up solutions are proposed to overcome these weaknesses.
For efficiency, sampling and approximation measures~\cite{omran2018scalable} are adopted to reduce time overheads of accurate rule evaluations; besides, many efficiency optimizations are proposed~\cite{Lajus2020FastAE} to speed up rule evaluation.
Nevertheless, the time-consuming candidate generation step is still inevitable.
For effectiveness, the partial completeness assumption (PCA) is proposed to debias the statistical estimation~\cite{galarraga2015fast}. 
However, this assumption only relieves the problem instead of totally solving it.
Therefore, it is still challenging for these improved solutions to balance between efficiency and effectiveness.
\subsection{Problem Analysis}
We reduce the above-mentioned problems to the \emph{inefficiency of rule generation} and \emph{ ineffectiveness of rule evaluation}.

{\bf The inefficiency of rule generation} results from the blind enumeration of all candidate rules.
These approaches only evaluate the quality of rules after the complete enumeration, incurring numerous unnecessary enumerations. A more reasonable solution is a stepwise  decision of the quality for a candidate rule immediately after its generation, avoiding meaningless enumeration.
The generation of a high-quality rule can be formalized as a \emph{sequential decision problem}. 
Given a head predicate, an intelligent rule miner should sequentially add a good predicate to the current unfinished candidate rule at each step.
At each time step, the rule miner makes the decision by evaluating the current state and possible actions, avoiding enumeration of all the possible rules.
For example, given the head predicate $nationality(x,y)$, the rule miner first considers the head predicate and adds a predicate $bornIn(x,z)$ to the body, forming an unfinished rule $bornIn(x,z) \land ... \rightarrow nationality(x,y)$.
In the next step, the rule miner then considers the current unfinished rule and decides to add another predicate $country(z,y)$ to the body, forming a complete rule $bornIn(x,z) \land country(z,y)\rightarrow nationality(x,y)$.
For sequential decision problems, reinforcement learning (RL) has been proven to be good at exploring and learning to make the best decisions~\cite{mnih2013playing,silver2017mastering,wang2021deep}.
Thus, RL provides a feasible solution for the rule generation problem in rule search.
A well-trained RL agent is able to reduce the enumeration space by avoiding bad decisions in the early stage of the rule generation.
{\bf The ineffectiveness of rule evaluation} could be attributed to the weakness of statistic measures. We argue that existing statistical measures are still not enough for effective evaluation, while distributed representations could be a significant complement to improve the evaluation.
Recently, the distributed representations, or embeddings, have been applied to many tasks since they contain rich latent information~\cite{bordes2013translating,socher2013reasoning}.
Many solutions utilize this latent information in rule mining to make an accurate evaluation of rule quality, achieving promising results	~\cite{yang2014embedding,neelakantan2015compositional,omran2018scalable,Ho2018LearningRF}.
Compared to explicit statistical measures, latent embedding measures have two obvious advantages.
First, it is able to capture latent information missing in explicit statistical measures.
Second, it is more tolerant to the noise of data, especially considering that the KGs are usually incomplete.
However, using embedding measures alone can hardly outperform statistical measures in general, and current major rule mining systems still use the statistical measures~\cite{Lajus2020FastAE}.
This motivates us to utilize them as an auxiliary signal to evaluate rules.

\subsection{Our Idea and Contribution}
{
	In this paper, we propose a generation-then-evaluation rule mining approach guided by reinforcement learning. The key of our solution is the evaluation of an enumerated rule (or an intermediate state as an incomplete rule). We trained an RL agent for rule generation which has a value function that evaluates the expectation of intermediate states (corresponding to enumerated rules). Hence, we propose a two-phased framework, which is illustrated in Figure~\ref{fig:framework}.
	In the first (offline) phase, we train an RL agent for rule generation from knowledge graphs. In the second (online) phase, we utilize the value function of the RL agent to guide the step-by-step rule generation in the rule search.
	Specifically, we use the intermediate state evaluation provided by the agent to prune bad decisions and enumerate good decisions with higher priorities. 
	Alternatively, we could directly generate rules with the agent without the second phase. 
	However, a generic RL agent can only generate a single optimal rule for a head predicate when the only optimal policy is employed, leading to a limited recall of high-quality rules. Hence, rule enumeration in a larger space is still inevitable, which motivates us to design the second phase.

	It is non-trivial to train an RL agent for rule generation.
	First, there are various options for the reward design.
	To define the reward, we choose embedding-based measures instead of statistical measures due to the following two reasons.
	For one thing, embedding-based measures are more noise-tolerant and can capture latent information, as mentioned above.
	For another, embedding-based measures, in general, can be efficiently calculated by vector operation, compared to counting-based statistical measures.
	Second, it is difficult to train the RL agent because immediate rewards are hard to evaluate when the rule is incomplete. To alleviate the training difficulty, we adopt a curriculum learning strategy~\cite{bengio2009curriculum} which trains the agent with settings from easy to difficult.
	The curriculum strategy helps the agent converge faster and perform better.
	In summary, our contributions are threefold.
	\begin{itemize}
		\item First, we propose a reinforcement learning solution training the RL agent with a curriculum learning strategy for rule generation. The trained RL agent is able to provide immediate evaluation for candidate rule generation.
		\item Second, we build a rule mining system by applying the value function of the RL agent to guide the search for high-quality rules efficiently.
		\item Third, we conduct experiments on several datasets and prove that our method achieves the state-of-the-art performance in terms of efficiency and effectiveness.
	\end{itemize}
}

\section{Overview}
In this section, we first introduce some preliminaries of knowledge graphs and rules. Then we formalize our rule mining problem. Finally, we introduce our solution framework.

\subsection{Preliminaries}
\paragraph{Knowledge Graph} A KG $K$ consists of two sets $K = (E, F)$, where $E$ is the set of entities and $F$ is the set of facts. 
A fact in a knowledge graph is represented as a triple $k=(s, P, o)$, which means that the subject entity $s$ is related to the object entity $o$ via predicate $P$. $P(s, o)$ is true if the fact $k=(s, P, o)$ exists in $K$.

\paragraph{Rules} Let $\Gamma = \{P\}$ denote the set of all the predicates.
In this paper, we focus on \emph{closed-path} (CP) rules because the syntax provides a balance between the expressive power of rules and the efficiency of mining. 
The form of a CP rule $r$ is
\begin{equation}
	P_1(x, z_1) \wedge P_2(z_1, z_2) \wedge ... \wedge  P_n(z_{n-1}, y) \rightarrow P_0(x, y).
	\label{eq:cp_rule}
\end{equation}
Here $x$, $y$ and $z_i$ are variables of entities, each $P_i(s, o)$ is called an atom, and $s$ and $o$ are called the subject and object argument for $P_i$ respectively.
$n$ is the length of the body, and the length of the rule is $n+1$.
The atom $P_0(x, y)$ is the head of $r$, and the body of $r$ is denoted as $B_r:=P_1(x, z_1) \wedge P_2(z_1, z_2) \wedge ... \wedge P_n(z_{n-1}, y)$.
The rule $r$ is \textit{closed-path} if the sequence of predicates in the rule body forms a path from the subject argument to the object argument of the head. 
Note that, inverse predicates are allowed in a CP rule, so rules like $bornIn^{-1}(Seattle, Bill\_Gates) \wedge nationality(Bill\_Gates, USA) \rightarrow country(Seattle, USA)$ are valid. 
Restricting the rules as closed-path rules is a standard formalism in the rule mining literature~\cite{galarraga2015fast,chen2016scalekb,omran2018scalable}.
For example, to make rules more expressive, AMIE+~\cite{galarraga2015fast} requires mined rules to be \emph{closed} and \emph{connected}; and in its experimental settings, literal values are eliminated; so, it basically only mines the closed-path rules in practice.

\paragraph{Rule Mining}
Given a head predicate $P_0$, the rule mining task studied in this paper is to mine rules in the form of Eq.~\eqref{eq:cp_rule} as many as possible from a KG $K$ within a certain time limit.

\subsection{Rule Evaluation}
Traditional rule mining approaches execute a search over the full rule space and evaluate the quality of each rule.
\subsubsection{Traditional Approaches} 
To assess the quality of rules, statistic-based measures are widely used in some mainstream approaches of rule learning~\cite{galarraga2015fast, omran2018scalable}.
Let $r$ be a CP rule of the form defined in Eq.~\ref{eq:cp_rule}. Then the \emph{support} of $r$ ($supp(r)$) is defined as the number of distinct entity pairs $(x, y)$ satisfying the rule $r$ in the KG:
\begin{equation}
	supp(r) = \mid \{(x, y): \exists z_1, ..., z_{n-1} :B_r \land P_0(x, y)\} \mid.
\end{equation}
The \emph{confidence} of $r$ ($conf(r)$) under the closed-world assumption (CWA) is the support normalized by the number of $(x, y)$ that satisfies the body constraint:
\begin{equation}
	conf(r) = \frac{supp(r)}{\mid\{(x, y):\exists z_1,...,z_{n-1}:B_r\}\mid}.
	\label{eq:conf}
\end{equation}
The \emph{head coverage} of $r$ ($hc(r)$)  is the support normalized by the number of $(x, y)$ that satisfies the head:
\begin{equation}
	hc(r) = \frac{supp(r)}{\mid\{(x, y):P_0(x, y)\}\mid}.
\end{equation}

\subsubsection{Embedding-based Approaches}
\label{sec:emb_rules}
Note that a path $p = \langle P_1, P_2, ..., P_n \rangle$ can be seen as a composite predicate between the starting entity and the ending entity, and $p$ is plausible if it is semantically similar to the head predicate $P_0$. 
For example, in a CP rule $bornIn(x, y) \wedge country(y, z) \rightarrow nationality(x, z)$, the path $p=\langle bornIn, country \rangle$ can be seen as a composite predicate about the entity pair $(x, z)$, and is semantically similar to the head predicate $P_0=nationality$.
Based on this intuition, \cite{yang2014embedding} defines a scoring function via an embedding model. We use the following rule of length 3 as an example to illustrate the method:
\begin{equation}
	P_1(x, z) \wedge P_2(z, y) \rightarrow P_0(x, y).
\end{equation}
Many KG embedding approaches (such as translation-based embedding model TransE~\cite{bordes2013translating}) can be used to derive an embedding for a path. In a CP rule, the embedding ($\mathbf{P_1} + \mathbf{P_2}$) of the predicate path $p = \langle P_1, P_2 \rangle$ in the body part should be similar to the embedding of its head predicate $\mathbf{P_0}$, that is, $\mathbf{P_1} + \mathbf{P_2} \approx \mathbf{P_0}$. Hence, the quality of a rule can be quantified by the similarity between $\mathbf{P_1} + \mathbf{P_2}$ and $\mathbf{P_0}$:
\begin{equation}
	score_1(r) = sim(\mathbf{P_1}+\mathbf{P_2}, \mathbf{P_0}).
	\label{eq:transe_sim}
\end{equation} 
Alternatively, the quality score could be also calculated as the bilinear form~\cite{socher2013reasoning} 
\begin{equation}
	score_2(r) = sim(\mathbf{P_1}\cdot \mathbf{P_2}, \mathbf{P_0}).
	\label{eq:bilinear_sim}
\end{equation} 
The similarity function $sim(\cdot, \cdot)$ has various alternative options, such as cosine similarity or Euclidean norm.
Anyone could be used in our solution without any significant influence on the results.

\begin{figure*}
	\centering
	\includegraphics[width=\columnwidth]{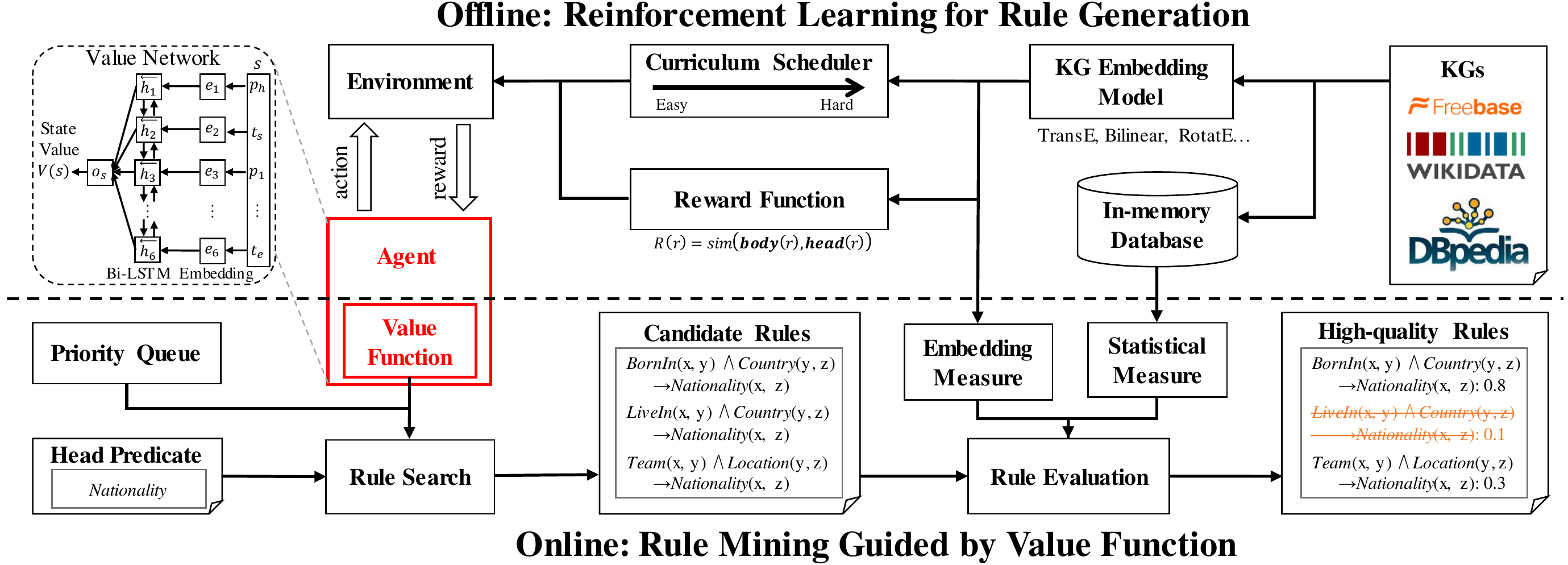}
	\caption{The framework of our solution. We first train an RL agent to generate high-quality rules with curriculum learning strategy and KG embedding reward. Then we use the value function of the trained agent to guide the search for candidate rules. We evaluate candidate rules with a hybrid measure combining explicit statistical and latent embedding measures.}
	\label{fig:framework}
\end{figure*}

\subsection{Solution Framework}

The framework of our approach is presented in Figure~\ref{fig:framework}. 
Our approach consists of two phases, including offline reinforcement learning for rule generation and online RL-guided rule mining.
In the offline phase, we perform reinforcement learning for rule generation.
We take the KG as the input, train the KG embeddings as a rule evaluation measure.
Then we use the embedding measure as the reward to train an RL agent for rule generation.
For policy learning, we adopt a temporal-difference value function approximation algorithm and a curriculum learning strategy.
After training, the RL agent is able to provide guidance during the rule generation.
The online procedure takes a head predicate as input and mines high-quality rules about it.
To mine the rules of all the predicates in the KG, we just enumerate all the predicates and execute the online procedure.
In the rule search procedure, each candidate rule is generated step-by-step and guided by the value function of the agent. 
Besides, we introduce a priority queue in order to control the search order.
For the candidate rules generated in the rule search procedure, we evaluate them with a hybrid measure combining the statistical and embedding measures.
In the rest of this paper, we first elaborate on our offline phase in Section~\ref{sec:offline} and then introduce the online rule mining algorithm guided by value function in Section~\ref{sec:online}.


\section{Reinforcement Learning for Rule Generation}
\label{sec:offline}
\begin{figure}
	\centering
	\includegraphics[width=\columnwidth]{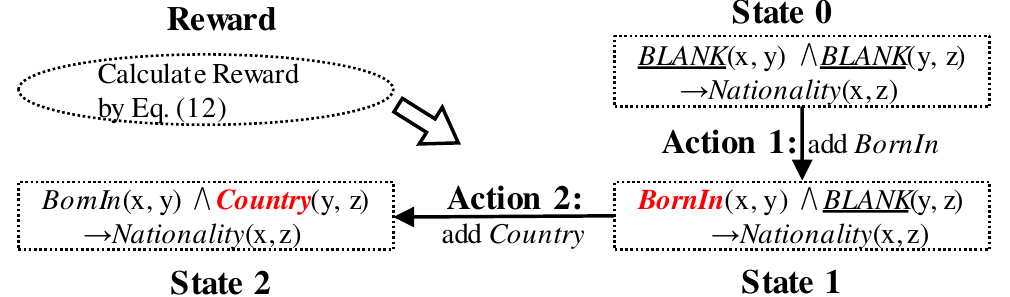}
	\caption{The example of rule generation formalized as MDP.}
	\label{fig:mdp}
\end{figure}

The generation of a high-quality rule is formalized as a sequential decision problem. 
Given a head predicate, we sequentially add a predicate to the current unfinished candidate rule at each step.
As shown in Figure~\ref{fig:mdp}, given the head predicate $nationality(x,y)$, we first add a predicate $bornIn(x,z)$ to the body, deriving an unfinished rule $bornIn(x,z) \land \cdots \rightarrow nationality(x,y)$.
Then we add another predicate $country(z,y)$ to the body, forming a complete rule $bornIn(x,z) \land country(z,y)\rightarrow nationality(x,y)$.
%
Therefore, the rule generation task can be modeled as a Markov Decision Process (MDP).
We design a reward function to measure the correctness of the current atoms of the rule, and train an agent to solve the MDP.
In the rest of this section, we first specify the details of the MDP components and then elaborate on the training of the agent.

\subsection{Markov Decision Process Components}
In general, the MDP is modeled as a tuple $(S, A, T, R)$, where $S = \{s\}$ is the space of all possible real-valued vector states; $A = \{a\}$ is the set of actions; since this is a deterministic MDP, transition $T: (S, A) \mapsto S$ refers to a  function that maps the domain of state and action to a deterministic state; the reward function $R: (S, A) \mapsto \mathbb{R}$ maps the domain of state and action to a real value.
To formalize the rule generation process as an MDP, we specify the definition of \emph{state} and \emph{action}, and describe the design of the \emph{reward} as follows.
\subsubsection{State and Action}
As shown in Figure~\ref{fig:mdp}, each state corresponds to a complete ($bornIn(x, y) \wedge country(y, z) \rightarrow nationality(x, z)$ for state 2) or an incomplete ($bornIn(x, y) \wedge BLANK(y, z) \rightarrow nationality(x, z)$ for state 1) rule.
At each decision point, the state consists of a sequence of body predicates and the head predicate
\begin{equation}
	P_1 \land P_2 \land ... \land P_n \rightarrow P_0.
\end{equation}
Thus, we define the state as a sequence of predicates in the rules, with additionally introduced special tokens,
\begin{equation}
    S = \{[\underbrace{P_0}_\text{head}, t_s, \underbrace{P_1, P_2, ..., P_n}_\text{body}]: P_0\in\Gamma \land P_i\in\Gamma\cup\{t_m\},  i=1,...,n, \forall n \in \mathbb{Z}^+\}
\label{eq:state}
\end{equation}
$t_s$ is a separation token to denote the separation between the head part and the body part.
We fix the body length $n$ when the state is initialized, and use a special mask token $t_m$ (like token ``BLANK" in Figure~\ref{fig:mdp}) to replace predicates that have not been determined in the incomplete rule.
The state is \textit{terminal} when it represents a complete rule, i.e., all the body predicates are not masked.

For convenience, we define a mapping $\phi$ from the space of states to that of rules.
We first define the space of all CP rules as $\Omega=\{P_1\land...\land P_n\rightarrow P_0:P_i\in\Gamma, i=0,1,...,n,\forall n\in\mathbb{Z}^+\}$.
For non-terminal states, they cannot be mapped to $\Omega$.
Then we use a special variable $\zeta$ to denote any incomplete rule.
Thus, the mapping $\phi: S \mapsto \Omega\cup \{\zeta\}$ is formally defined as follows.
For a state $s=[P_0, t_s, P_1, P_2, ..., P_n]\in S$,


\begin{equation}
	\begin{split}
		\phi(s) =
		\begin{cases} 
			P_1\land...\land P_n\rightarrow P_0,  & \mbox{if }~P_i \in \Gamma, \forall i=1, ..., n \\
			\zeta,  & \mbox{otherwise}
		\end{cases}.
	\end{split}
\end{equation}
The function $\phi$ can also be used as an identifier for terminal states, i.e. the state $s$ is terminal iff $\phi(s)\in\Omega$.

In the initial state, the head predicate $P_0$ is assigned to a random predicate and the undetermined body predicates are masked.
The body predicates can be masked entirely (when we only know the head predicate) or partially (when we know the head predicate and some of the body predicates).

The action for each state is to materialize a masked body atom with a specific predicate in KG. Based on the state defined as above, 
each action in the set is formalized as a tuple
\begin{equation}
	A = \{(u, v): u = 1,...,n \land v\in \Gamma\},
\end{equation}
where $u$ is the index of the body predicate in state sequence and $v$ is a predicate.
The transition of $T(s, a)$ for an action $a=(u,v)$ is to set the $u$-th body atom of $s$ to predicate $v\in \Gamma$.

Compared to approaches for common sequential tasks such as text generation, our approach has several special designs.
First, we fix the length of the sequence before the generation.
It is because generating predicates in a rule is sensitive to the length.
Besides, it will be convenient for our curriculum learning strategy to control the difficulty of the generation process by initializing the state with different lengths.
Second, we do not generate the sequence in a pre-defined order (e.g. left to right). 
We let the agent to decide both the order and content. In this way, the agent can generate more confident predicates first.


\subsubsection{Reward}
The reward for an action is expected to encourage the agent to generate a good rule.
The quality of a generated rule can be only assessed when the complete rule is generated, which means we can only provide rewards for those actions leading to terminal states.
Thus, we define the reward for an action $a$ on the state $s$ as follows

\begin{equation}
	\begin{split}
		R(s, a)=
		\begin{cases} 
			\rho(\phi(s')),  & \mbox{if }~\phi(s')\in \Omega \\
			0,  & \mbox{otherwise}
		\end{cases},
	\end{split}
\end{equation}
where $s'=T(s, a)$ is the next state.
When the next state $s'$ is a terminal state, which means it represents a complete rule, we provide the reward by an evaluation function $\rho$.
As mentioned before, the evaluation of a CP rule can be realized by any KG embedding models that satisfy composition patterns~\cite{sun2019rotate} such as TransE~\cite{bordes2013translating} and bilinear function~\cite{socher2013reasoning}.
In this paper, we adopt the TransE model to realize the score function, although our framework can be generalized to other methods.
Specifically, for a rule $r$ with the form specified in Eq.~\eqref{eq:cp_rule}, we calculate the embedding score $\rho(\cdot)$ as follows
\begin{equation}
	\rho(r) = sigmoid(\eta - ||\mathbf {P_0}-\Sigma_{1 \le i \le n}\mathbf {P_i}||_1)
\end{equation}
where $\eta$ is a hyper-parameter, and $\mathbf{P_i}$ is the TransE embedding of the predicate $P_i$.

The function $\rho$ is an instantiation of Eq.~\eqref{eq:transe_sim}, i.e. measuring the similarity between the embedding of the rule body, $\Sigma_{1 \le i \le n}\mathbf {P_i}$, and that of the head, $\mathbf {P_0}$.
Since the training objective of TransE is to minimize the distance of embeddings~\cite{bordes2013translating}, we use the distance in the training phase to formalize the similarity in order to directly utilize the expression power of the embeddings.
The sigmoid function and the hyper-parameter $\eta$ are used to normalize the similarity score from 0 to 1.
We train the embedding with the re-implementated version of TransE~\cite{sun2019rotate}, and use the same hyper-parameter $\eta$ from training phase (the hyper-parameter is $\gamma$ in their paper).

\subsection{Agent Training}
In this section, we elaborate on the details in the training of the RL agent, including the model of the agent, the policy learning algorithm, and the curriculum learning strategy. 

\subsubsection{Model} 
The state is a sequence of predicates (Eq.~\eqref{eq:state}) and each predicate is sequentially dependant on the context predicates.
Thus, as shown in the top-left corner in Figure~\ref{fig:framework}, we encode the state with an LSTM~\cite{hochreiter1997long} network since it has been proven to be good at capturing sequential information.
The network for state encoding is formalized as the composition of three functions 
\begin{equation}
	\mathbf{o_s}=G(s)=h(g(f(s))). 
\end{equation}
First, it looks up the sequence of tokens through an embedding layer, $f(\cdot)$. 
Second, it encodes them with stacked bidirectional LSTM layers, $g(\cdot)$.
Then, it obtains the state representation $\mathbf{o_s}\in \mathbb{R}^{d_o}$ by averaging over all the LSTM output logits along the sequence dimension, $h(\cdot)$. 
Finally, we obtain the state value by projecting the state encoding into a real number 
\begin{equation}
	V(s; \theta) = sigmoid(\mathbf {W}\cdot\mathbf{o_s} + b).
\end{equation}
The value network $V(s; \theta)$ estimates the expected cumulative reward under current state $s$ with trainable parameters $\theta$ including $\mathbf{W}\in\mathbb{R}^{1\times d_o}$, $b\in \mathbb{R}$ and parameters in $G$.
\subsubsection{Policy Learning} 
	As for policy learning, we adopt a temporal-difference value function approximation algorithm to train the neural network.
	The details of our algorithm are illustrated in Algorithm~\ref{alg:value_iteration}.
{ \clhrev
The algorithm iteratively executes two steps in each episode, i.e. the simulation step (lines 3-9) and the learning step (lines 10-11).
In the simulation step, we obtain an initial state (line 3) and sample a trajectory of states and actions in the MDP (lines 4-9).
The generation of the trajectory is a Markov chain Monte Carlo (MCMC) sampling process~\cite{andrieu2003introduction} or a Markov sampling process~\cite{jiang2020svm,zou2021learning}, which is guided by the $\epsilon$-greedy policy~\cite{sutton2018reinforcement} derived from the value function (line 5).
The $\epsilon$-greedy policy is a stochastic policy that chooses a random action with $\epsilon$ probability and chooses the action achieving the largest state value with $1-\epsilon$ probability.
The $\epsilon$ parameter is to control the trade-off between the exploitation of the learned policy (the one achieving optimal value) and the exploration of other options.
We linearly decrease the $\epsilon$ value from $0.95$ to $0.05$ as the training proceeds and the value function is well trained.
We store the MDP tuple into the replay memory $M$ (line 7) at each decision making point.
In the learning step, we sample a batch of tuples from the memory (line 10) to train the value network (line 11).
}
	Specifically, given each MDP tuple $(s,a,s',R(s,a))$ in the memory, we update the value function through the Q-value $Q(\cdot,\cdot)$ as follows:
	\begin{equation}
		\begin{split}
			Q(s, a) & \leftarrow R(s, a) + \gamma \max_{a'}{V(T(s',a');\theta)}, \\
			\theta & \leftarrow \theta -\alpha\nabla||V(s'; \theta) - Q(s, a)||_1,
		\end{split}
		\label{eq:value_update}
	\end{equation}
	where $\gamma$ denotes the discount factor and $\alpha$ is the learning rate.

\begin{algorithm}
	\caption{Temporal-difference Value Function Approximation}
	\label{alg:value_iteration}
	\begin{algorithmic}[1]
		\Require{reward function $R$, transition function $T$, parameters of Algorithm~\ref{alg:curriculum}}
		\Ensure{Trained value network $V(\cdot;\theta)$}
		\State Initialize replay memory $M=\emptyset$
		\For{each episode}
    		\State Initialize state $s$ with Algorithm~\ref{alg:curriculum}
    		\While {$s$ is not terminal}
    		    \State Choose action $a$ for state $s$ using $\epsilon$-greedy policy derived from $V$
    		    \State Take action $a$, observe reward $R(s,a)$ and next state $T(s,a)$
    		    \State Push tuple $(s,a,T(s,a),R(s,a))$ into the memory $M$
    		    \State Proceed with the next state $s\leftarrow T(s,a)$
    		\EndWhile
    		\State Sample a batch $B$ from memory $M$
    		\State Update the value network $V(\cdot;\theta)$ with Eq.~\eqref{eq:value_update}
		\EndFor
	\end{algorithmic}
\end{algorithm}

	For the policy learning, two common alternatives are the deep Q-network (DQN) algorithm~\cite{mnih2013playing} and the policy gradient (PG) algorithms~\cite{silver2014deterministic}.
	However, the action space of our MDP is parametric~\cite{gauci2018horizon}, which depends on the current state.
	It is impractical to apply a traditional DQN or PG algorithm.
	Existing solutions represent actions along with states, as a set of features and produce Q-values or the policy for each state-action pair.
	However, the action space is large and good actions for each state are sparse when the relation vocabulary is large.
	The model can hardly be trained sufficiently in this setting. 
	Thus, we adopt such a value function approximation algorithm to train the neural network.
	This algorithm can be viewed as a variant of the Q-learning algorithm.
	Unlike Q-learning algorithm that updates the Q-network with many parameters for the calculation of each action value, we only calculate the state value function in our model, which reduces the parameters of sparse action embedding in the deep model.


\subsubsection{Curriculum Learning}
\label{sec:cl}
Since it is hard to judge whether a rule is good or bad when it is incomplete, there are no immediate rewards for all the actions in our RL generation procedure. 
Training such an agent is difficult at the beginning because the lack of immediate reward leads to ineffective updates on value function $V(s; \theta)$ according to Eq.~\eqref{eq:value_update}.

\begin{algorithm}
	\caption{Initialization with curriculum setting}
	\label{alg:curriculum}
	\begin{algorithmic}[1]
		
		\Require{a length distribution $\Phi$, a probability $q$, the set of predicates $\Gamma$, and a set of seed rules $\Omega$ }
		\Ensure{an initial state $s$}
		\State Generate a random real number $p \in [0, 1]$ 
		\If{$p > q$}
		\State Sample a rule $r$ of length $l$ from $\Omega$ 
		\State Construct state $s$ from rule $r$ 
		\State Uniformly sample an integer $m$ from $[1, l]$ 
		\State Uniformly mask $m$ body predicates in state $s$ 
		\Else
		\State Sample a length $l$ from distribution $\Phi$ 
		\State Uniformly sample a predicate $P_0$ from $\Gamma$ 
		\State Construct state $s$ with head predicate $P_0$ and $l$ masked body predicates 
		\EndIf
	\end{algorithmic}
\end{algorithm}

We adopt a curriculum learning strategy~\cite{bengio2009curriculum} to relieve this difficulty.
The difficulty of a generation process is attributed to two factors, the \emph{length of the rule} and the \emph{number of masked predicates}.
Rule length can be easily controlled in our special design of the state.
However, we cannot initialize the generation process with a specific number of masked predicates because we do not have any ground truth rules. 
Therefore, we first obtain some seed rules (body length smaller than 4) with high embedding scores by randomly sampling.
Then we use these rules as seeds in the initial steps of our curriculum learning setting.
Specifically, our curriculum setting consists of several stages, through which the initialized generation processes become harder.
In each stage, we initialize the generation process in the random procedure in Algorithm~\ref{alg:curriculum}.
We first randomly decide whether to fetch a rule from the seed rules with a probability $q$.
If the rule is not from the seeds, we randomly select a predicate as the head and construct an empty rule with the length sampled from the distribution $\Phi$.
The difficulty of the initialized generation process can be controlled by $\Phi$ and $q$.
We improve the probabilities of long rules in $\Phi$ and $q$ through stages to improve the generation difficulty.
Specifications of hyper-parameters $\Phi$ and $q$ will be given in the experiment section.

%


\section{Rule Mining Guided by Value Function}
\label{sec:online}
Since the rule mining task requires finding high-quality rules as many as possible, the exploration of the entire rule search space is inevitable.
We use the value function $V(\cdot)$ from the trained RL agent as a heuristic to guide the rule search procedure to find high-quality rules efficiently and effectively.
The value function provides evaluation for intermediate states by estimating the expected reward, which could serve as guidance for an effective search strategy. In general, an effective rule search strategy is expected to have two capabilities:  (1) pruning the states of low values correctly and (2) exploring the states of high values first. 

\begin{table}
	\centering
	\begin{tabular}{lll}  
		\toprule
		Notation & Type & Description \\
		\midrule
		$K$ & set & Source knowledge graph \\
		$V$ & function & Value function of RL agent \\
		$P_0$ & predicate & Head predicate of mined rules \\
		$L$ & integer & Length of mined rules ($\ge 2$) \\
		$B$ & integer & Batch size for value computation ($\ge 1$) \\
		$minC$ & float & Minimum confidence of mined rules \\
		$minH$ & float & Minimum head coverage of mined rules \\
		$minV$ & float & Minimum value for pruning \\
		\bottomrule
	\end{tabular}
	\caption{Specifications for inputs of Algorithm~\ref{alg:rule_search}.}
	\label{tab:input_args}
\end{table}

\begin{algorithm}
	\caption{Rule mining guided by Value Function} 
	\label{alg:rule_search} 
	\begin{algorithmic}[1]
		
		\Require{$K$, $P_0$, $L$, $B$, $V$, $minC$, $minH$, and $minV$ }
		\Ensure{a set of CP rules $Rules$ }
		\State Initialize an empty max heap $H :=  \emptyset$ 
		\State Initialize a rule $r_0$ with empty body and head $P_0$
		\State $Q := \{r_0\}$ 
		\State $Rules := \emptyset$ 
		\While{$H \ne \emptyset$ or $Q \ne \emptyset$}
    		\If{$H = \emptyset$ or $|Q|\ge B$} // Maintaining Phase
        		\For{$r \in Q$}
            		\If{$V(r) \ge minV$}
                		\State Push the tuple $(r, V(r))$ into the Heap $H$
            		\EndIf
        		\EndFor
        		\State $Q := \emptyset$
    		\EndIf
		    \If{$H \ne \emptyset$} // Exploring Phase
        		\State Pop the tuple $(r, v)$ out of the Heap $H$ 
        		\If{$length(r) < L$}
        		    \State $Q := Q \cup refine(r)$ 
        		\Else{\ \textbf{if} $S(r)\ge minC \land hc(r) \ge minH$}
        		    \State $Rules := Rules \cup \{r\}$
        		\EndIf
    		\EndIf
		\EndWhile
	\end{algorithmic}
\end{algorithm}

The details of our rule search strategy are described in Algorithm~\ref{alg:rule_search}.
The inputs for our algorithm are shown in Table~\ref{tab:input_args}.
The outputs are the mined rules with length $L$.
Our algorithm adopts a breadth-first search framework.
We use a queue $Q$ to store intermediate states to be explored and a max heap to control the search order and filter out bad choices.
Each iteration consists of a \emph{maintaining phase} (line 6) and an \emph{exploring phase} (line 14).
In the maintaining phase (lines 6-13), we take a batch of candidate intermediate states from the queue $Q$ and push them into the max heap $H$.
Specifically, we first take a batch of intermediate states from $Q$ (line 7). 
Then we compute the value of this batch of states and filter out bad choices (line 8).
Last, we push the rest of them into the maximum heap $H$ (line 9).
In the exploring phase (lines 14-20), explore the state with the maximum value from the heap $H$.
Specifically, we first select the state with the largest value from the heap $H$ (line 12).
If the state is terminal (line 15), the rule is evaluated by our rule evaluation measure $S$ which will be elaborated in the next section.
Otherwise, the algorithm uses the refinement operator $refine$ (line 14) to derive new states by enumerating all possible atoms that can be appendedd to the body.
The list $Q$ serves as a buffer to support the batch computation of the value function (lines 7-9).
The algorithm finishes when running out of rules or the time limit is reached.

For rule evaluation, we adopt the hybrid evaluation measure~\cite{Ho2018LearningRF} via adding the embedding score $\rho$ as a complement to the original statistical measure $\psi$ as follows.
\begin{equation}
	S(r) = \lambda\cdot \psi(r) + (1-\lambda)\cdot \rho(r), 
	\label{eq:score_function}
\end{equation}
where $\psi$ is the statistical confidence as defined in Eq.~\eqref{eq:conf} or other definition such as PCA confidence~\cite{Galrraga2013AMIEAR}, and $\lambda \in [0, 1]$ is the hyper-parameter trading off the influence between the statistical measure $\psi$ and the latent embedding measure $\rho$.

Although we use the value function learned from embedding reward to guide the candidate rule generation, it is still necessary to incorporate embedding-based measures into the rule evaluation for two reasons.
First, the advantage of the value function is to provide immediate evaluation for incomplete rules, and it is hard to outperform embedding measures on complete rule evaluation.
Second, quantified measures need to be provided for the rule evaluation to select high-quality rules.

\section{Experiments}
In this section, we evaluate the effectiveness of the proposed approach.
We conduct experiments to evaluate two aspects.
\begin{itemize}
	\item Performance of agent. The RL agent is trained to get higher rewards of embedding, which has been proved to be effective on rule mining. Hence achieving high performance on the reward is necessary for the results of rule mining task.
	\item Performance of rule mining. We conduct experiments to prove the efficiency and effectiveness of our rule miming approach guided by the value function. 
\end{itemize}

\subsection{Experimental Setup}
\begin{table}
	\centering
	\begin{tabular}{p{2cm}>{\centering\arraybackslash}p{2cm}>{\centering\arraybackslash}p{2cm}>{\centering\arraybackslash}p{2cm}}  
		\toprule
		KG & \# Facts & \# Entities & \# Predicates \\
		\midrule
		WN18RR & 86.8K & 40.6K & 11 \\
		FB15K-237 & 310K & 14.5K & 237 \\
		Wikidata & 8.40M & 4.00M & 430 \\
		DBpedia 3.8 & 11.0M & 2.20M & 650 \\
		\bottomrule
	\end{tabular}
	\caption{Knowledge graph specifications.}
	\label{tab:datasets}
\end{table}
To prove the scalability of our approach, we conduct experiments on two full-size KGs and two benchmark datasets specified in Table~\ref{tab:datasets}, where the last two have been often used in rule mining~\cite{galarraga2015fast,chen2016scalekb}. FB15K-237 is a refined set of FB15K~\cite{bordes2013translating}, which is obtained from Freebase and used in the rule mining task~\cite{yang2014embedding}.

We implement our method in Python3.7.6. We use Tensorflow 1.14.0 and Keras 2.2.4 to implement the deep learning part.
Our experiments were conducted on a computer with Intel(R) Core(TM) i7-8700K CPU at 3.70GHz, GPU of GTX 1080Ti, and 32GB of RAM, running Ubuntu Linux 18.04.

\begin{table}
	\centering
	\begin{tabular}{ccccccc}  
		\toprule
		\multirow{2}{*}{Stage} & \multicolumn{5}{c}{probabilities of length in $\Phi$}  & \multirow{2}{*}{$q$} \\
		\cmidrule(lr){2-6}
		& 2 & 3 & 4 & 5 & 6 & \\
		\midrule
		0 & 0.25 & 0.25 & 0.50 & 0.00 & 0.00 & 0.0 \\
		\midrule
		1 & 0.17 & 0.33 & 0.50 & 0.00 & 0.00 & 0.3 \\
		\midrule
		2 & 0.15 & 0.20 & 0.25 & 0.40 & 0.00 & 0.6 \\
		\midrule
		3 & 0.10 & 0.15 & 0.20 & 0.25 & 0.30 & 0.8 \\
		\bottomrule
	\end{tabular}
	\caption{The 4-stage curriculum setting.}
	\label{tab:curriculum_setting}
\end{table}

\textbf{KG Embedding.} For embedding training, we use the TransE model realized by \cite{sun2019rotate} with a negative sample size of 256, dimension of 1000, and $\eta$ of $24.0$.
For some datasets, the entity sets are too large for the embedding training.
We filter out entities with less than 9 relations in DBpedia3.8, resulting in a subset of 120,805 entities, 434 relations, and 736,803 facts. 
For Wikidata, we filter out entities with less than 7 relations, resulting in a subset of 119,484 entities, 275 relations, and 581,691 facts.
Although about a third of relations are pruned in both two datasets, the facts to which remaining relations are related both account for over 90\% of the whole set.
Hence the pruned relations are long-tailed with few instances.
Despite that most of the facts are pruned when training the embedding, we use all the related facts for rule mining.


\textbf{Curriculum.} Table~\ref{tab:curriculum_setting} specifies the curriculum settings of the 4 stages.
As the stage lasts, the process of rule generation becomes more difficult.

{\clh
	\textbf{RL.} The value network has a token embedding layer with the dimension of 256, LSTM hidden units of 512, and $d_o=1024$.
	We find that more complex neural networks with more layers or hidden units will not improve the performance of reward in the reinforced training.
	In our experiments, the agent is trained 50K, 100K, 100K, and 150K for each stage respectively.
	We adopt the RMSprop optimizer with the learning rate $\alpha=0.001$. 
	We use the replay memory of 10,000 and the training batch size of 128.
	We set the discount factor $\gamma=0.99$.
}
\subsection{Performance of RL Agent}
In this section, we conduct experiments to prove the effectiveness of our value function approximation algorithm for agent training by comparing it with some other common options.
\subsubsection{Evaluation of Policy Learning}

\begin{figure}
	\subcaptionbox{stage 0\label{fig:state_sf1}}{\includegraphics[width=0.45\columnwidth]{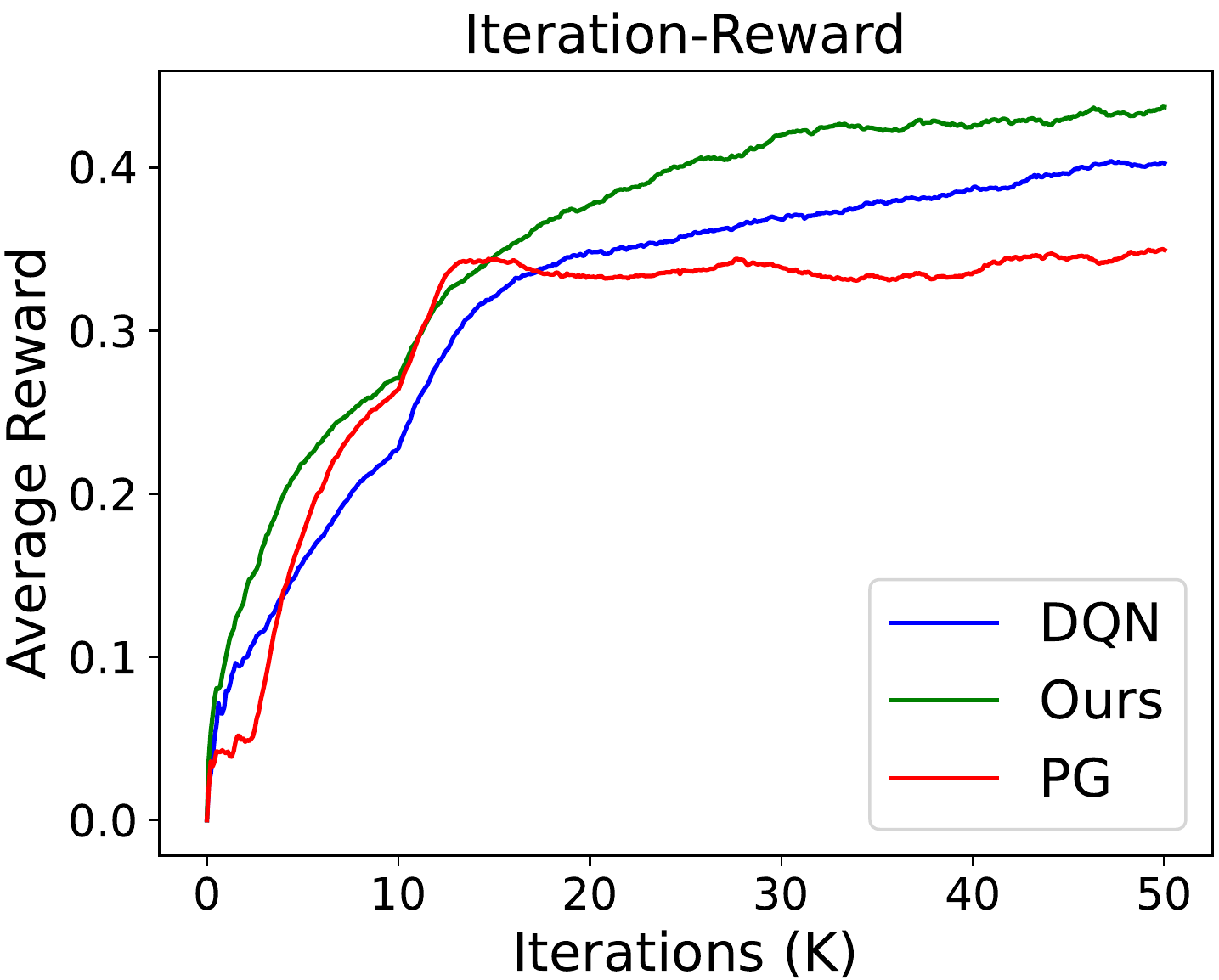}}
	\hspace{1em}
	\subcaptionbox{stage 1\label{fig:state_sf2}}{\includegraphics[width=0.45\columnwidth]{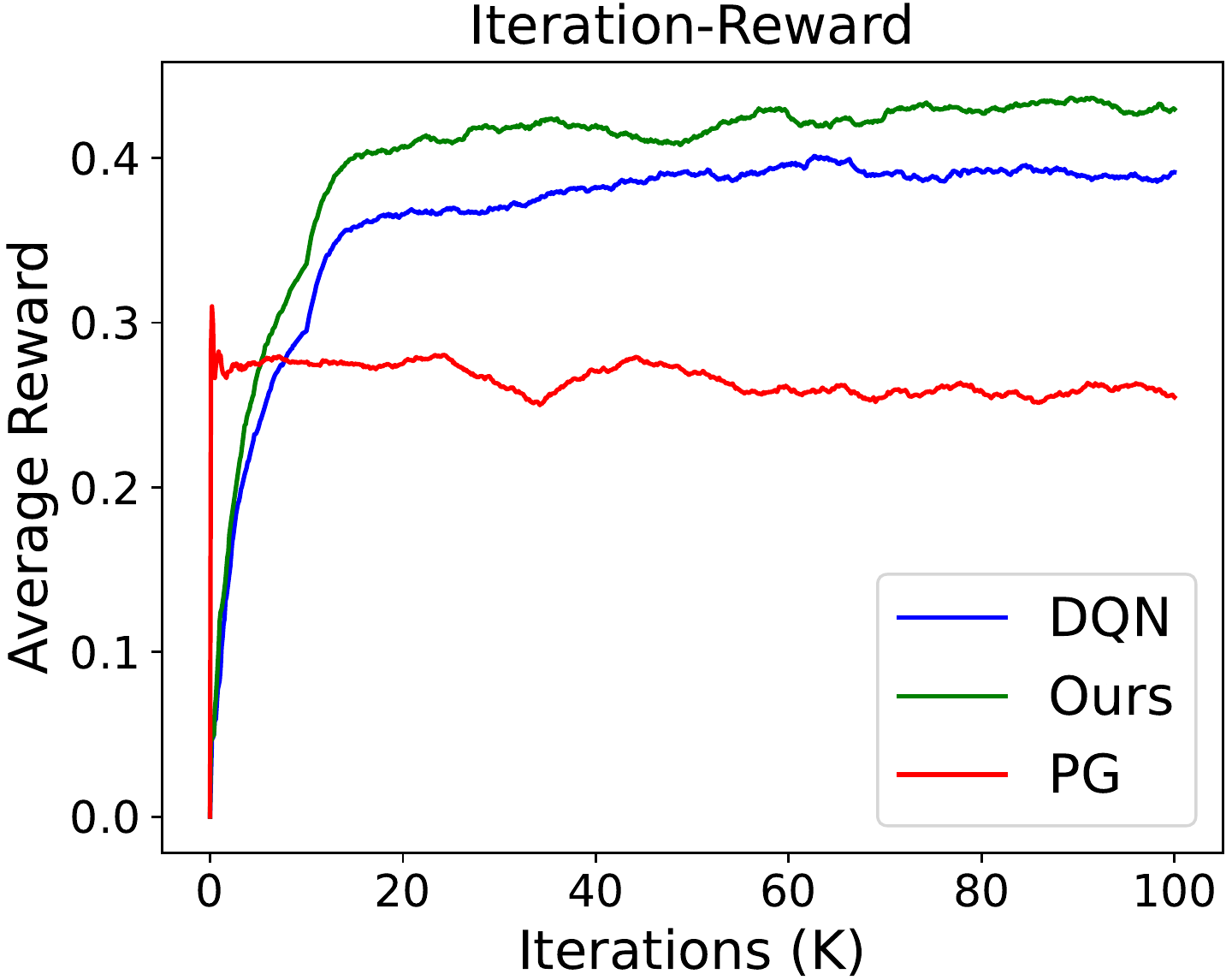}}
	\caption{Comparison of our algorithm with other baselines in the curriculum stage 0 and 1. Our algorithm has the best performance.}
	\label{fig:state}
\end{figure}

To alleviate the problem of the parametric-action DQN with large action space, we only train a value network and produce the policy by the value of the next states.
To prove the effectiveness of this design, we compare it with other options.
The comparisons are as follows.
\begin{itemize}
	\item DQN. We adopt the existing solution of using the action embedding concatenated with state embedding to calculating Q-values~\cite{gauci2018horizon}. We use the action embedding of dimension $1000$ and the state embedding generated by the LSTM encoding of the end token $t_e$.
	\item Policy Gradients. In this baseline, we adopt the policy gradient algorithm to optimize the agent. We use the classic PG algorithm, REINFORCE~\cite{sutton2018reinforcement}, to optimize the same network settings as the baseline above.
\end{itemize}
We compare the baselines with our methods in the curriculum settings of stage 0 and 1.
The results in Figure~\ref{fig:state} show that the our algorithm has the best performance against other competitors.
In addition, the performance gain by the curriculum learning strategy is not very significant for PG algorithm compared to DQN and ours.

\subsubsection{Evaluation of Curriculum Learning}
\begin{figure}
	\subcaptionbox{stage 2\label{fig:curriculum_sf1}}{\includegraphics[width=0.45\columnwidth]{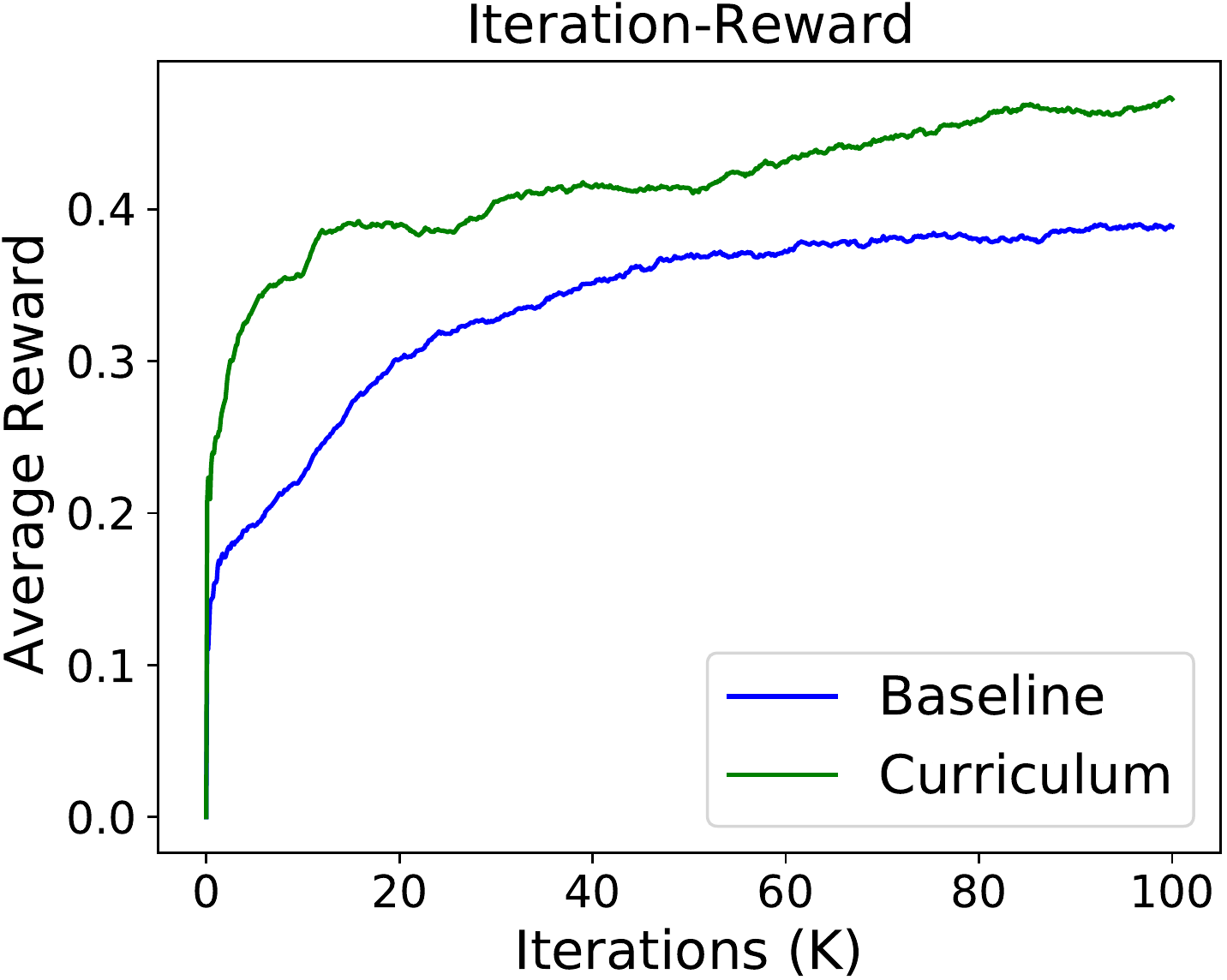}}
	\hspace{1em}
	\subcaptionbox{stage 3\label{fig:curriculum_sf2}}{\includegraphics[width=0.45\columnwidth]{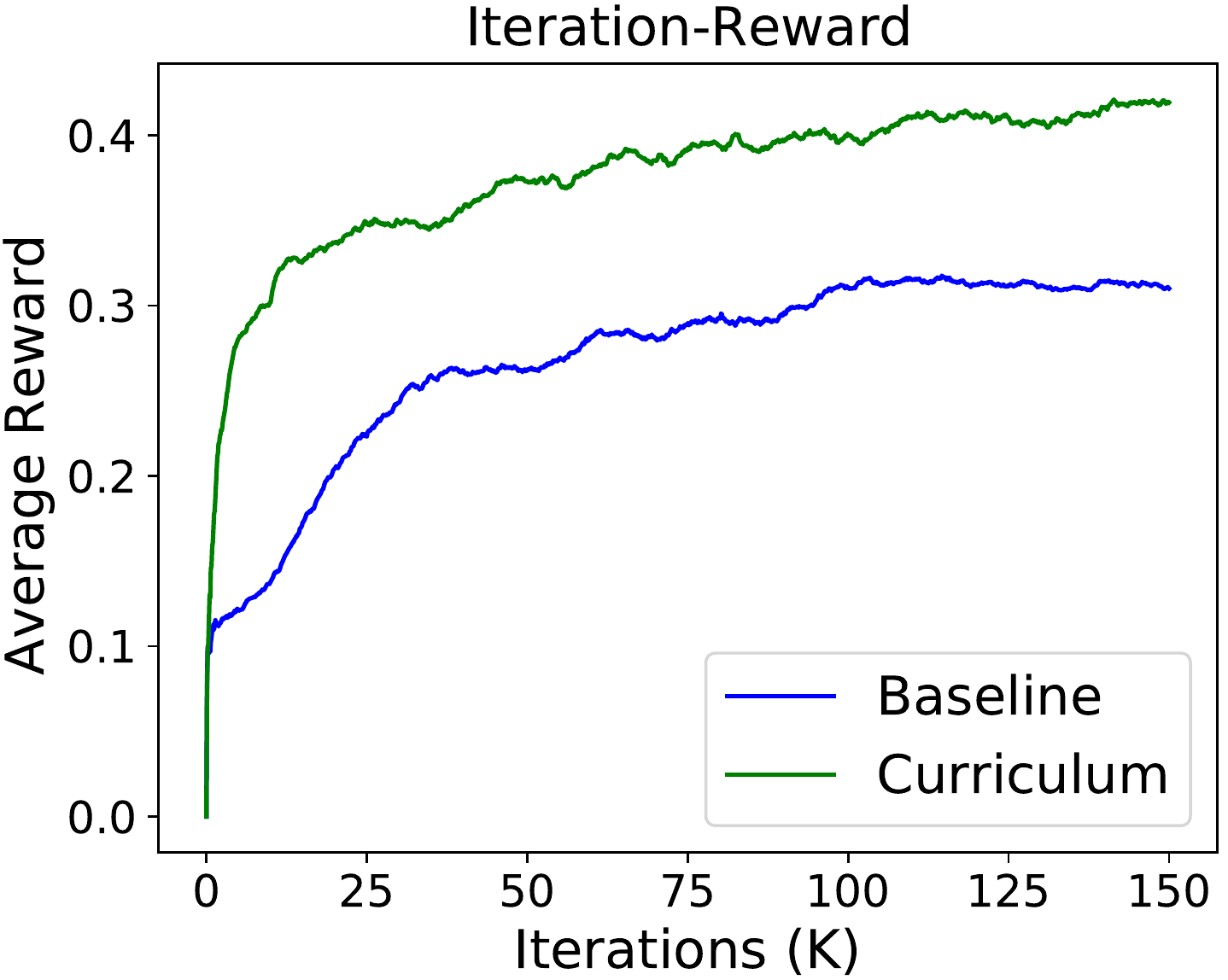}}
	\caption{Results of curriculum learning in stage 2 and 3. Our curriculum learning strategy significantly improves the performance of the RL agent on reward-iteration curves.}
	\label{fig:curriculum}
\end{figure}

To train the agent better without immediate reward, we a adopt curriculum learning strategy to train the agent from easy to difficult.
In this section, we conduct experiments to prove that our curriculum learning strategy is effective.
We compare the agent trained by our curriculum learning strategy to the baseline in the last two stages.
The baseline has the same setting except for the previous stages of training.
The results in Figure~\ref{fig:curriculum} show that our curriculum learning strategy significantly improves the performance of the RL agent on reward-iteration curves.
Moreover, the margin of the performance becomes larger when the task gets harder.
This means the effectiveness of curriculum learning strategy becomes more prominent when the task gets harder.
In addition, the performance of the baseline is close to convergence.
That is to say, the performance margin cannot be narrowed by extending the training time of the baseline.
In conclusion, the optimization provided by curriculum learning strategy achieves not only faster convergence but also better performance.

\subsection{Performance of Rule Mining}
This set of experiments aims to evaluate the performance of our RL-guided rule mining approach.
The evaluation involves two aspects, the efficiency, whether our approach mines more rules than others, and the effectiveness, whether the rules mined by our approach are of high quality.
We conduct experiments in this section on the four datasets, WN18RR, FB15K-237, DBpedia 3.8, and Wikidata.
Note that the advantages of our approach are highlighted in the datasets with large relation space since we focus on reducing the rule search space.
Therefore, we do not include YAGO2s~\cite{suchanek2007yago} since it only contains 37 relations and does not have the problem of large rule space, and only use WN18RR for link prediction experiments.
The thresholds for rule mining are $minC=0.1, minH=0.01, minV=0.0001$. 
The batch size $B$ of our algorithm is 128.
We execute our algorithm with $L$ varying from small to large and the time limit is usually reached when $L=5$.


\subsubsection{Evaluation of Efficiency}
We conduct experiments to show our approach has better performance in terms of the quantity of mined high-quality rules in a certain time limit.
Although it takes several hours to train the KG embedding and RL agent, they are offline procedures and negligible when amortized to several hundreds of predicates in KG.
Thus, we exclude the offline training time in this experiment.

We compare the efficiency of our method with two rule miners AMIE+~\cite{galarraga2015fast} and RLvLR~\cite{omran2018scalable}.
The former is a classic rule mining system and the latter is an improvement based on the former one.
Although many optimizations have been proposed based on AMIE+~\cite{Lajus2020FastAE}, these optimizations are scalable to all the top-down rule mining methods including ours.
We do not consider those optimizations in our experiments and use AMIE+ as the baseline because we only need to prove the superiority of our rule generation strategy guided by the RL agent.
For the reproduction reason, we re-implement the RLvLR with our trained embeddings and build a stronger baseline than the original paper~\footnote{Our implementation of RLvLR achieves a much better performance than the original paper in following link prediction experiments.}.
Optimizations in AMIE 3 are also applicable in our rule mining framework.
In this experiment, we do not compare the efficiency of our method with rule miners such as RuDik~\cite{Ortona2018RobustDO} and AnyBURL~\cite{meilicke2020reinforced} because they generate rules with different manners.
For example, AnyBURL generates rules by sampling paths in the KG, and the measures such as support and confidence are approximately calculated by sampling, which is not comparable with our accurate rule evaluation.
In addition, since we only evaluate the efficiency of rules in this experiment, we use the confidence based on closed-world assumption as the rule evaluation measure for all the methods.

\paragraph{Quantity of Mined Rules}
We selected all the predicates as head predicates for FB15K-237, and 30 popular predicates\footnote{We rank predicates in descend order of their number of instances in the KG, and choose the top 30 for our experiments.} for Wikidata and DBpedia. 
We set a 1 hour limit for each head predicate in FB15K-237 and 5 hour limit for each in Wikidata and DBpedia.
Table~\ref{tab:quantity_rule_mining} shows the average numbers of quality rules (\#Rules, conf$\ge$ 0.1 and hc$\ge$ 0.01) and the average number of high-quality rules (\#Q-Rules, conf$\ge$ 0.7) for selected head predicates.

\begin{table}
	\centering
	\setlength{\tabcolsep}{3.5mm}{
		\begin{tabular}{l|l|r|r}  
			\hline
			Dataset & Method & \#Rules & \#Q-Rules \\
			\hline
			\multirow{3}{*}{FB15K-237} & AMIE+ & 500.09 & 89.65 \\
			& RLvLR & 1923.35 & 312.99 \\
			& Ours & \textbf{3095.38} & \textbf{544.21} \\
			\hline
			\multirow{3}{*}{DBpedia 3.8} & AMIE+ & 10.20 & 1.63\\
			& RLvLR & 34.90 & 1.87 \\
			& Ours & \textbf{66.60} & \textbf{8.50}  \\		
			\hline
			\multirow{3}{*}{Wikidata} & AMIE+ & 4.63 & 1.50 \\
			& RLvLR & 15.67 & 4.00 \\
			& Ours & \textbf{24.07} & \textbf{8.00} \\
			\hline
	\end{tabular}}
	\caption{Comparison of the average number of mined rules. The number of rules we mine is significantly larger than that of AMIE+ and RLvLR.}
	\label{tab:quantity_rule_mining}
\end{table}

Results from Table~\ref{tab:quantity_rule_mining} show that our method significantly outperforms baseline methods in terms of mining both the quality and high-quality rules.
The superiority of our method is more obvious in mining high-quality rules. 
The numbers of quality rules mined by our method on DBpedia 3.8 and Wikidata are nearly 2 times larger than those mined by RLvLR.
The performance of our method is less prominent on the FB15K-237 dataset.
According to our analysis, it is a subset of Freebase and has much fewer instances than a regular KG, which leads to less time cost for evaluating each rule.
As the time cost for rule evaluation drops, the amortized overheads of our rule search optimization are higher in the whole rule mining procedure.
Nevertheless, the superiority of our method is more highlighted as the KG size grows larger.

\begin{figure}
	\includegraphics[width=0.9\columnwidth]{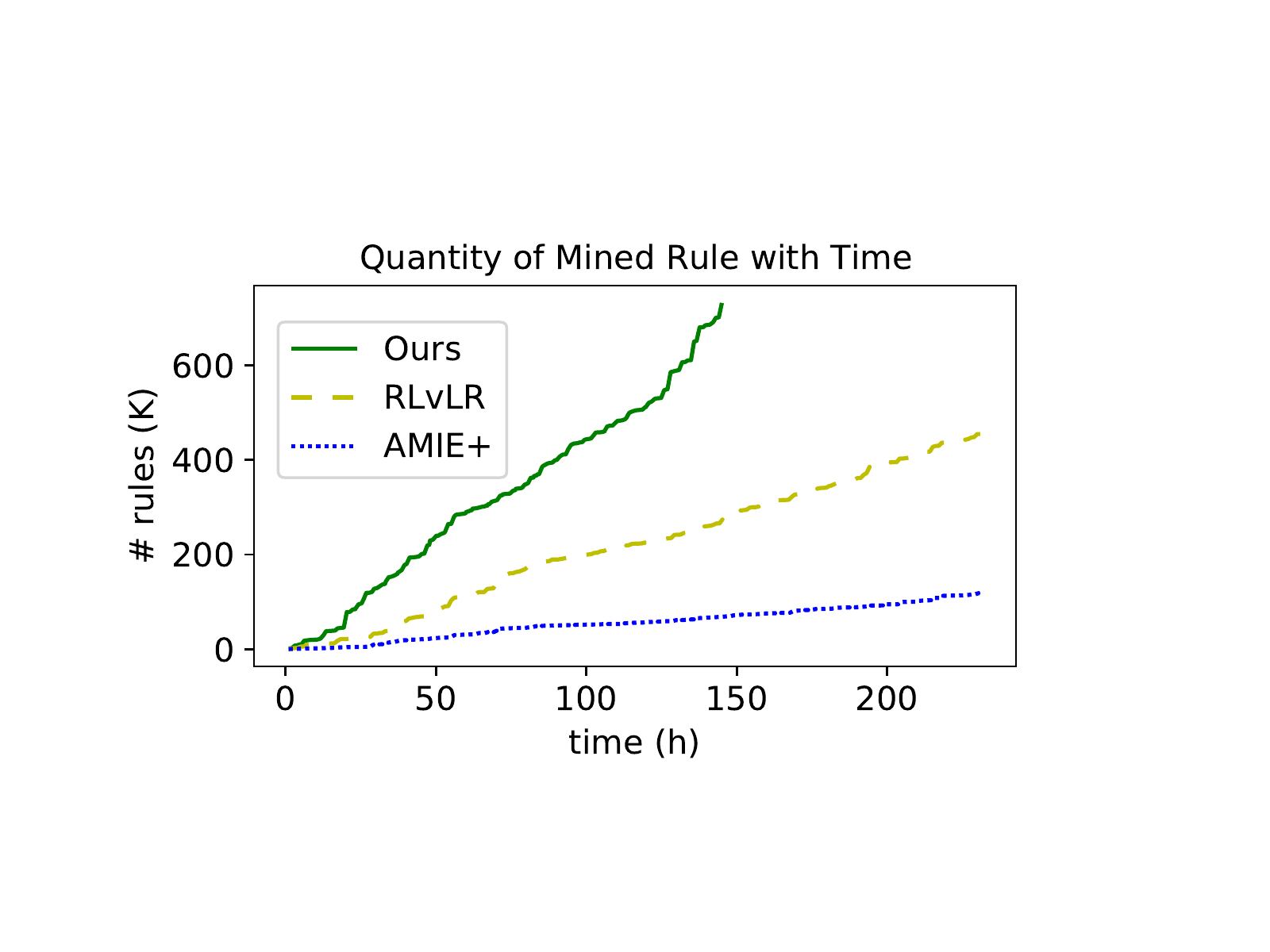}
	\caption{The number of mined rules with time. Our method mines more rules with less time.}
	\label{fig:rc_time}
\end{figure}

{\clhrev
Although we set time limits for all the comparative methods, we find our method stops early for many predicates within the pre-defined length $L=5$.
Thus, we plot the mined rule quantity with time for all comparative methods on the FB15K-237 dataset to further show the efficiency of our method.
As shown in Figure~\ref{fig:rc_time}, our method spends around only 150 hours on the mining of 237 predicates in FB15K-237 and mines more rules than comparative methods in over 200 hours.
}

\paragraph{Quantity of Predicted Facts}
We also conduct experiments to evaluate the predictive power of mined rules.
This evaluation is necessary because a method mining more rules does not mean those mined rules have more predictive power since they might just cover the existing facts.
To obtain the confidence degree (CD) of a fact inferred by rules, we adopt the score function from~\cite{galarraga2015fast,omran2018scalable}, by aggregating the confidence of all the rules inferring the facts in a Noisy-OR manner. 
The intuition is that facts inferred by more rules should have a higher confidence degree.
Formally, for a fact $k=(s, P, o)$ and the set of rules $Rules$ that can infer $k$ from the given KG, the CD of $k$ is defined as follows:
\begin{equation}
	CD(k) = 1-\prod_{r\in Rules}{(1-S(r))}, 
	\label{eq:confidence_degree}
\end{equation}
where $S(r)$ is the score function for rule $r$. 

\begin{table}
	\centering
	\setlength{\tabcolsep}{3.5mm}{
		\begin{tabular}{l|l|r|r|r}  
			\hline
			Dataset & Method & \#Facts & \#QFacts & {\clhrev Time (s)}\\
			\hline
			\multirow{3}{*}{DBpedia 3.8} & AMIE+ & 491323 & 245758 & {\clhrev 15  }\\
			& RLvLR & 359457 & 171358 & {\clhrev 185 }\\
			& Ours & \textbf{629680} & \textbf{297253} & {\clhrev 179 } \\		
			\hline
			\multirow{3}{*}{Wikidata} & AMIE+ & 432469 & 277186  & {\clhrev 9 }\\
			& RLvLR & 484192 & 305350 & {\clhrev 1088 }\\
			& Ours & \textbf{740842} & \textbf{486732} & {\clhrev 1093 }\\
			\hline
	\end{tabular}}
	\caption{Comparison of the predictive power. \#QFacts and \#Facts are the counts of predicted facts that are in the held-out sets. Our method is significantly better than baselines.}
	\label{tab:predictive_power}
\end{table}

For the estimation of the predictive power of mined rules, we follow the setting of \cite{omran2018scalable} and conduct experiments on DBpedia and Wikidata.
We eliminated from each dataset 30\% of its facts involving the head predicates as the held-out sets and checked how many facts in the held-out sets can be predicted by applying mined rules on the remaining facts.
Unlike \cite{omran2018scalable}, we only count the predicted facts in the held-out sets because the prediction of existing facts does not show the predictive power of rules.
We do not use FB15K-237 in this experiment because it is not a full-scale KG.
Table~\ref{tab:predictive_power} shows the numbers of predicted facts (\#Facts) and those predictions with $CD \ge 0.7$ (\#QFacts). 
From the results, we can see that our method outperforms others in both datasets about the predictive power.
In our experimental settings, RLvLR mines more rules than AMIE+ on DBpedia while those rules have less predictive power.
Note that, these two metrics only count the predicted facts that are in the held-out set, which can be regarded as the ``recall" of the predicted facts.
Thus, the ratio of ``\#QFacts" among ``\#Facts" is not an evaluation metric for the performance of algorithms.
For the quality of predicted facts, we will evaluate in the next section with the precision metric.

{\clhrev 
In addition, we also present the comparison of prediction time for the predictive experiment.
Since the prediction procedure is shared among all the methods, the factor leading to the different prediction time is the number of mined rules.
As shown in the Table~\ref{tab:predictive_power}, our method requires large prediction time due to the large quantity of rules we mine.
However, AMIE+ consumes significantly less prediction time, which is not proportional to the number of mined rules (our method mines around 5 times more rules than AMIE+ while requiring over 100 times more prediction time in Wikidata).
This is because AMIE+ mines rules from short to long, and it usually stops at short rules ($L \le 3$) when the time limit is reached.
On the contrary, most rules our method mines are long rules ($L \ge 4$).
Since the time complexity for applying a rule in KG is exponential to its length, the prediction time of our method is significantly larger than AMIE+.
As for RLvLR, it mines long rules as ours but ignores the sufficient discovery of short rules, thus it exhibit poor performance when predicting facts in large KG.
}


\subsubsection{Evaluation of Rule Quality}
The second set of experiments aim to evaluate the quality of mined rules of our approach.
Previous works~\cite{galarraga2015fast,Ortona2018RobustDO} evaluated the quality of mined rules by examining the new facts they discover.
Follow that criterion, we first evaluate the discovered new facts of mined rules in DBpedia and Wikidata, and then conduct experiments of link prediction task on FB15K-237 dataset.
For a fair comparison, we use the CWA confidence as the statistical rule evaluation measure for all the methods.
We tune the hyper-parameter $\lambda=0.9$ via link prediction performance on the FB15K-237 validation set and apply it to all experiments. 
\begin{figure}
	\centering
	\subcaptionbox{DBpedia \label{fig:prec_DB}}{\includegraphics[width=0.9\columnwidth]{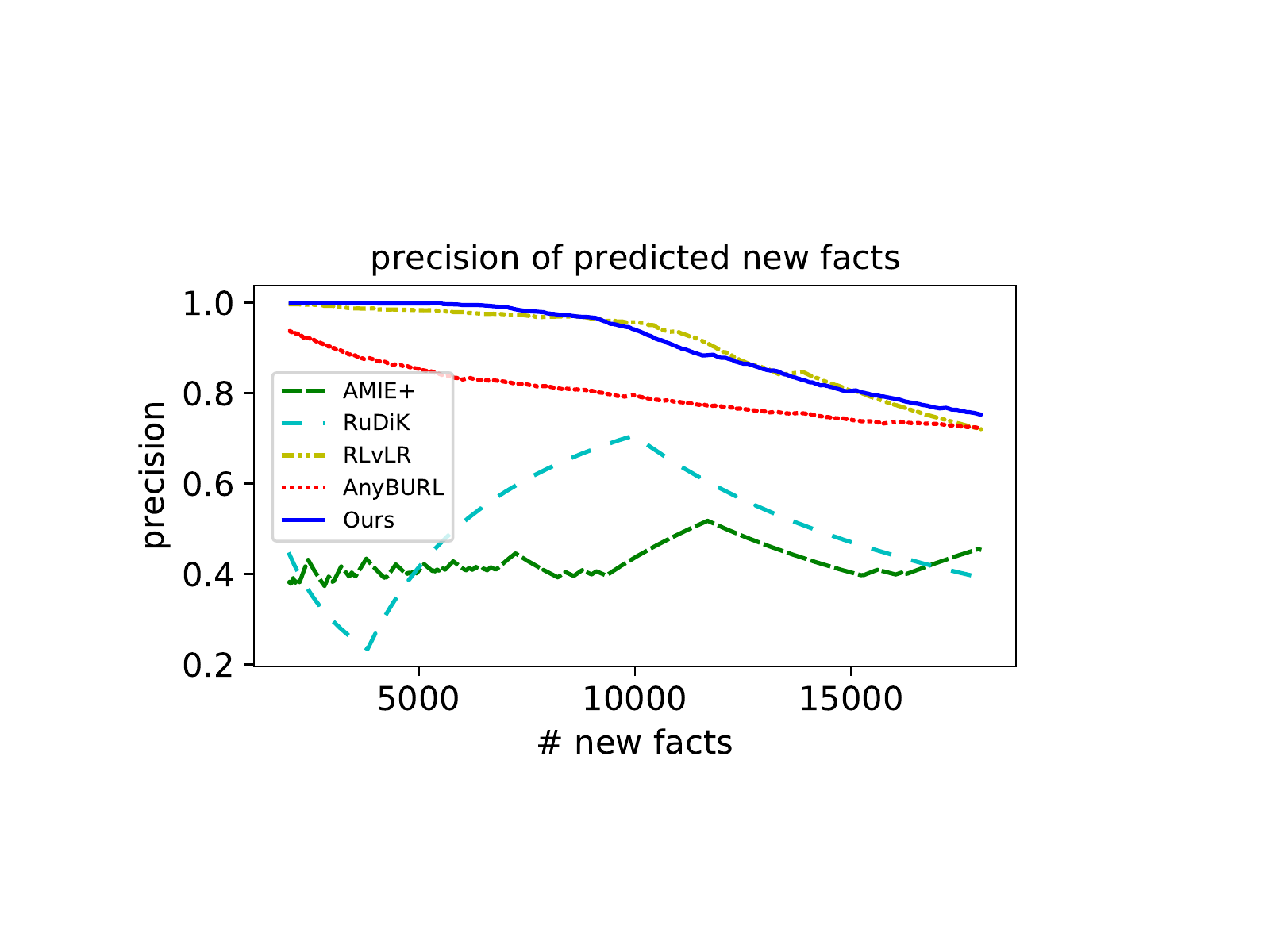}}
	\hspace{1em}
	\subcaptionbox{Wikidata \label{fig:prec_Wiki}}{\includegraphics[width=0.9\columnwidth]{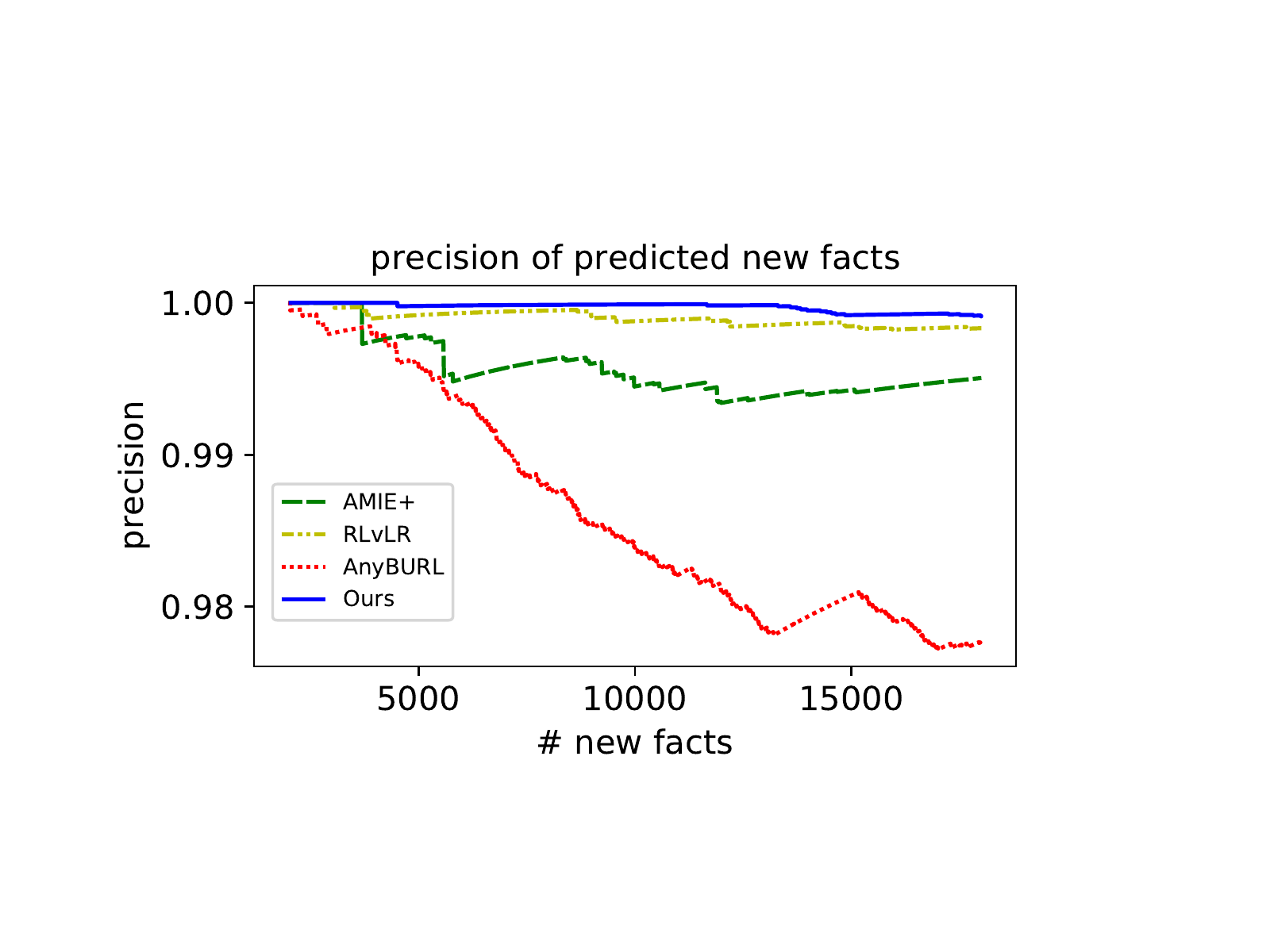}}
	\caption{Precision comparison of new facts predicted by rules. The facts are ranked by the descending order of the predicted scores.}
	\label{fig:precision}
\end{figure}

\paragraph{Precision of Predicted Facts}
First, we evaluate the quality of mined rules by examining the precision of new facts they discover for large knowledge bases, such as DBpedia and Wikidata.
Applying the previous setting, we randomly select 30\% of facts from each dataset as the held-out set and check the precision of predicted facts by applying rule mining methods on the remaining facts.
We use the same 30 predicates and the same settings as previous experiments. 
Due to various limitations, many rule mining methods only employ on benchmark datasets instead of full-size KGs.
Thus, in this set of experiments, we only compare our methods against state-of-the-art scalable rule mining systems, such as AMIE+~\cite{galarraga2015fast}, RLvLR~\cite{omran2018scalable,omran2019embedding}, RuDik~\cite{Ortona2018RobustDO}, and AnyBURL~\cite{meilicke2020reinforced}.
Furthermore, we only mine closed-path rules for those baselines.
Thus, we only use positive rules for RuDik and constraint them to be \emph{closed} and \emph{connect} and without literal values, which is the same as the experiment setting of AMIE+\cite{galarraga2015fast}.
We rank the predicted facts by descending order of $CD$ scores from Eq.~\eqref{eq:confidence_degree} for each method, and examine their precision (whether the predicted fact is in the held-out set or not).
The results in Figure~\ref{fig:prec_DB} and Figure~\ref{fig:prec_Wiki} show that the performance of our method is superior to other competitors.
In the DBpedia dataset, RLvLR has comparable performance, though our method has the best performance when the number of predicted facts is smaller than 10000 or larger than 15000.
In the setting of Wikidata, our method outperforms competitors marginally but steadily. 
The performance of RuDik is not comparable to these methods, so we exclude it from this figure to make the difference of presented methods prominent.

\begin{table}
	\centering
	\setlength{\tabcolsep}{3.5mm}{
		\begin{tabular}{l|l|cc|cc}  
			\hline
			\multirow{2}{*}{Category} & \multirow{2}{*}{Method} & \multicolumn{2}{c|}{FB15K-237} & \multicolumn{2}{c}{ WN18RR} \\
			&  & MRR & Hits@10 & MRR & Hits@10 \\
			\hline
			\multirow{5}{*}{\makecell{Embedding\\-based}} 
			& TransE~\cite{bordes2013translating} & 29.4 & 46.5 & 22.6 & 50.1\\
			& DistMult~\cite{yang2014embedding} & 24.1 & 41.9 & 43 & 49\\
			& ComplEx-N3~\cite{lacroix2018canonical} & 37 & 56 & 48 & 57 \\
			& RotatE~\cite{sun2019rotate} & 33.8 & 53.3 & 47.6 & 57.1\\
			& TuckER~\cite{balavzevic2019tucker} & 35.8 & 54.4 & 47.0 & 52.6 \\
			\hline
			\multirow{2}{*}{\makecell{Rule-aware}} 
			& RNNLogic+~\cite{qu2020rnnlogic} & 34.9 & 53.3 & \textbf{51.3} & \textbf{59.7} \\
			& RARL~\cite{hou2021rule} & \textbf{55.7} & \textbf{63.4} & 46.9 & 53.3 \\
			\hline
			\multirow{8}{*}{\makecell{Rule\\Learning}} 
			& AMIE+~\cite{Galrraga2013AMIEAR} & - & 40.9 & - & 38.8\\
			& Neural LP~\cite{yang2017differentiable} & 23.7 & 36.1 & 38.1 & 40.8 \\
			& RLvLR*~\cite{omran2018scalable} & 34.2 & 49.9 & 44.7 & 53.0\\
			& DRUM~\cite{sadeghian2019drum} & 34.3 & 51.6 & 48.6 & \textbf{58.6}\\
			& NLIL~\cite{yang2019learn} & 25 & 32.4 & - & -\\
			& GPFL~\cite{gu2020towards} & 32.2 & 50.4 & 44.9 & 55.2\\
			& AnyBURL~\cite{meilicke2020reinforced} & 34.5 & 52.0 & \textbf{49.9} & 57.2\\
			& Ours & \textbf{35.4} & \textbf{52.4} & 49.3 & 57.9\\ 
			\hline
	\end{tabular}}
	\caption{Comparison of link prediction task with other methods. Although many methods achieve better performance in the benchmark datasets, we mainly focus on the comparison with \emph{explainable} and \emph{inductive} methods in rule learning literature, among which our method shows prominent performance. [*] means we rerun the methods with our implementation.}
	\label{tab:link_prediction}
\end{table}

\paragraph{Link Prediction}
We also evaluate the quality of mined rules by their performance on the link prediction task in benchmark datasets.
Given a KG, the task of link prediction is to identify for each predicate $P$ and each entity $s$, an entity $o$ such that $(s, P, o)$ is in the KG; or alternatively, to identify for each predicate $P$ and each entity $o$, an entity $s$ such that $(s, P, o)$ is in the KG.
FB15K-237 and WN18RR are two challenging and widely used benchmarks for link prediction.
We include the methods in rule learning literature~\cite{Galrraga2013AMIEAR,yang2017differentiable,omran2018scalable,sadeghian2019drum,yang2019learn,gu2020towards,meilicke2020reinforced}, and others~\cite{bordes2013translating,yang2014embedding,lacroix2018canonical,sun2019rotate,balavzevic2019tucker,qu2020rnnlogic,hou2021rule} for comparison.
Since we focus on evaluating the quality of the rules, we mainly compare with methods in the rule mining literature.
These baselines usually satisfy two conditions.
First, they must produce explicit and explainable rules for prediction. 
There are many embedding-based~\cite{sun2019rotate,balavzevic2019tucker} or graph neural network-based~\cite{schlichtkrull2018modeling} methods with high performance but low interpretability.
Second, they must produce inductive prediction (i.e. not rely on the head or tail entities).
Many methods produce high performance with dependence on the information of entities learned in the training phase~\cite{qu2020rnnlogic,hou2021rule}.
Although those methods show better performance on the benchmark, their ability of generalization for large-scaled KGs is limited compared to the inductive methods.
The experimental results are summarized in Table~\ref{tab:link_prediction}.
Our method shows prominent performance among rule learning methods on FB15K-237 dataset, while it only presents competitive results on WN18RR dataset.
This is because our method focuses on pruning the search space of rule discovery procedure.
For a complex dataset like FB15K-237 (237 relations), the search space is relatively large and the advantages of our method are prominent.
However, for a relatively simple dataset like WN18RR (11 relations), the search space is relatively small and the utility of our method is limited.


\subsubsection{Analysis of Rule Evaluation Measure}	
\begin{figure}
	\includegraphics[width=0.9\columnwidth]{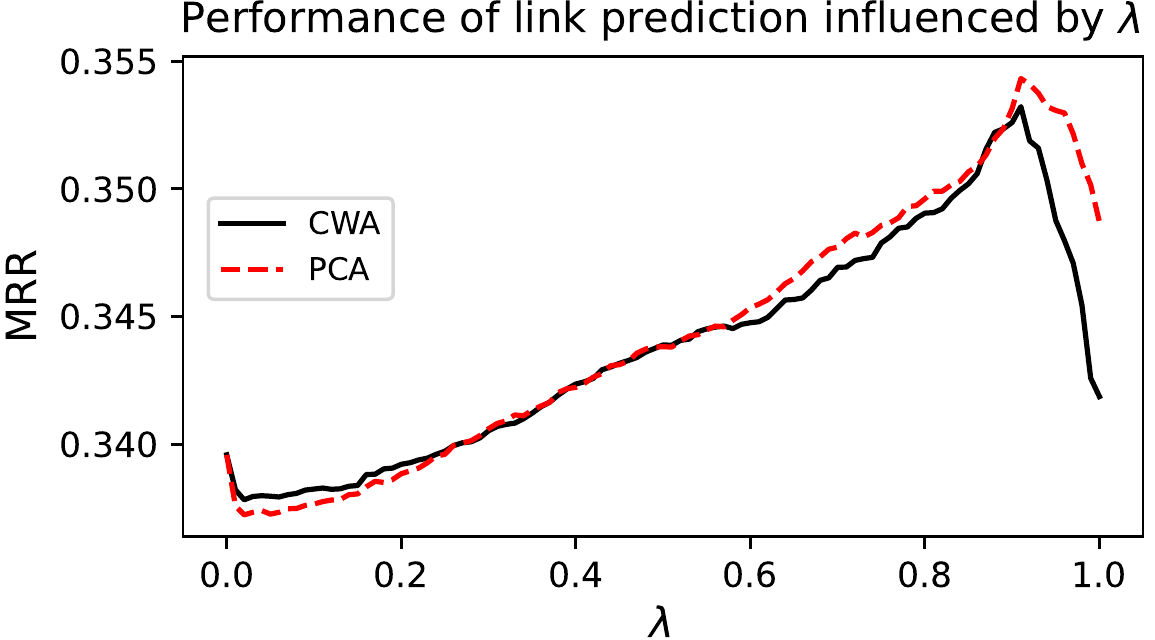}
	\caption{The influence of measure trade-off parameter $\lambda$ on link prediction performance on FB15K-237 test set.}
	\label{fig:lambda_mrr}
\end{figure}

For a deep dive into the rule evaluation measure, we conduct experiments to analyze how statistical measures and embedding measures influence the rule evaluation.
We compare two different statistical measures, CWA confidence, and PCA confidence.
To analyze the influence of embedding measures, we change the hyper-parameter $\lambda$ and observe how the results change.
When $\lambda=1$, the rule evaluation only relies on the statistical measure; when $\lambda=0$, it only relies on the embedding measure.
The results are shown in Figure~\ref{fig:lambda_mrr}.
We can see that the best ensemble performance is achieved when $\lambda$ is around $0.9$.
Besides, PCA confidence has better performance than CWA confidence although the difference is marginal. 
The results also explain that even the value network is learned from the embedding reward, it is still necessary to incorporate embedding measures in the rule evaluation for better performance.

\subsubsection{Discussion: Why the RL method is better.}
Theoretically, the key distinction between our RL method and traditional rule mining methods is that we model the rule generation problem as a sequential decision making problem instead of brutal-force enumeration.
This distinction provides our methods advantage to make decisions during the step-by-step generation of the rules.
We propose the Algorithm~\ref{alg:rule_search} for rule search, where we generate the rules the guidance of trained value function.
Thus, the effectiveness of our RL method results from the pruning strategy based on the value function of the RL agent.
During the rule search procedure, the value function provides evaluation for the intermediate states and guide the search algorithm to explore the states with higher scores.
Since the search space of candidate rules is infinite, this strategy can improve the efficiency and effectiveness of rule mining procedure within a certain time limit.
Although the overall performance shows the effectiveness of our method, it is a little obscure that how the guidance from the value function helps the rule mining procedure.
We realize the key to the effectiveness of our method is how well the evaluation provided by the value function corresponds to the genuine rule quality.
Thus, we randomly select 10000 intermediate states in the rule search process of our experiments on FB15K-237 dataset, and calculate the relevance between the evaluation provided by the value function ($V$) and the corresponding ratio of quality rules (conf$\ge$ 0.1) produced from the state ($QR$).
The two variables $V$ and $QR$ shows high Pearson correlation of $0.728$.
This reflects the effectiveness of the guidance provided by the value function.

{\clh
	\subsubsection{Case Study and Error Analysis}
	
	In this section, we present some case studies of our rule mining approach.
	First, we present some good cases of quality rules mined by our system.
	Then, we examine some error cases and analyze the reasons.
	
	We observe many informative rules learned by our method but neither RLvLR nor AMIE+ could learn them. 
	Here are some examples in Table~\ref{tab:case_study}.
	Since short rules are easily enumerated by traditional rule miners, we specially present some informative long rules to emphasize the effectiveness of our approach.
	Because of the large search space, these long rules are very hard for traditional rule mining methods to discover.
	Our method is able to mine these quality rules very efficiently due to the strong pruning strategy provided by the value function.
	
	\begin{table}
		\centering
		\setlength{\tabcolsep}{3.5mm}{
			\begin{tabular}{l|l}  
				\hline
				Head & Body \\
				\hline
				\multirow{4}{*}{$nationality$} & $bornIn(x, z_1) \land marriedIn(z_2, z_1) $ \\
				&  $\land graduateIn(z_2, z_3) \land country(z_3, y)$\\
				\cline{2-2}
				& $graduateIn(x, z_1) \land sportsTeam(z_1, z_2) $ \\
				&  $\land teamLocation(z_2, z_3) \land country(z_3, y)$\\
				\hline
				\multirow{2}{*}{$filmLanguage$} & $location(x, z_1) \land releaseRegion(z_2, z_1)  $ \\
				&  $\land storyBy(z_2, z_3) \land personLanguage(z_3, y) $\\
				\hline
				\multirow{2}{*}{$hometown$} & $associatedBand(z_1,x) \land hometown(z_1, z_2)  $ \\
				&  $\land owner(z_3, z_2) \land location(z_3, y) $\\
				\hline
				\multirow{2}{*}{$filmCountry$} & $actIn(z_1,x) \land produce(z_1, z_2)  $ \\
				&  $\land crewMember(z_2, z_3) \land nationality(z_3, y) $\\
				\hline
				
		\end{tabular}}
		\caption{Examples of rules mined by our approach. The variables of heads ($x$ and $y$) are omitted.}
		\label{tab:case_study}
	\end{table}
	
	However, there are also some bad cases produced by our method.
	A major problem of our system is that it mines many the trivial rules with no actual predictive power.
	For example, our method mines rule $awardNomination(e_1, s) \land nominatedFor(e_1, e_2)  \land honoredFor(o, e_2) \rightarrow awardCeremony(s, o)$ from Freebase.
	This rule means if $e_1$ is nominated by the award $s$ for his/her work $e_2$, and the work $e_2$ is honored for the ceremony $o$, then the award $s$ is likely to be presented in the ceremony $o$.
	An instance of $(s, e_1, e_2, o)$ is (\emph{Primetime Emmy Award for Outstanding Lead Actor in a Drama Series}, \emph{Bryan Cranston}, \emph{Breaking Bad}, \emph{60th Primetime Emmy Awards}).
	Although this type of rules makes sense, it cannot be used to predict new facts in the knowledge graphs because the fact of the head predicate $awardCeremony$ is much easier to be acquired than those of body predicates.
	This problem cannot be eliminated by the current rule evaluation measure because this rule has high score in both statistical and embedding measures.
	It is an interesting topic to mine rules that has actual predictive power, which we will leave it as the future work.
}
\section{Related Work}
We survey the related work from three aspects.

\textbf{Rule mining.} Traditional methods of rule learning mine high-quality rules by calculating statistical measures such as supports and confidence~\cite{galarraga2015fast,Ortona2018RobustDO,Lajus2020FastAE}.
Even though many pruning strategies and new measures such as partial completeness assumption confidence are proposed, the lack of scalability and the limitations of the statistical measures are still huge challenges for those methods.
Restricting the rules as closed-path rules is a standard formalism in the rule mining literature~\cite{chen2016scalekb,omran2018scalable,omran2019embedding}.
Some adopt the embedding from representation learning to score rules~\cite{yang2014embedding,omran2018scalable,omran2019embedding}, which provide extra information for evaluating rules but are still not very efficient compared to some major rule miners.
Our solution combines statistical and embedding information and provides scalability for the mining of long rules.

{\clh
	\textbf{Reinforcement learning.} Previous works have shown that reinforcement learning methods are good at exploring and making the best decisions in large search space~\cite{mnih2013playing,silver2017mastering,wang2021deep}.
	It has been widely applied in KG-related tasks~\cite{Fang2019JointEL}. 
	The most related tasks to rule mining is KG  reasoning~\cite{wang2020adrl,qu2020rnnlogic,hou2021rule}.
	Those applications utilize RL to solve the optimization problems such as finding an optimal path in the KG.
	The main difference in applying RL to the rule mining task is that we need to discover as many quality rules as possible instead of finding the optimal one.
	Recently RL has been applied to the rule mining task~\cite{meilicke2020reinforced} in a bottom-up manner.
	Our work utilizes RL to mine rules in a top-down manner and takes advantage of the latent information provided by distributed representation to achieve better performance.
}

\textbf{Curriculum learning.} \cite{bengio2009curriculum} provided a good overview of curriculum learning in a task-specific way.
Recently more curriculum strategies have been developed for reinforcement learning, such as teacher-guided curriculum~\cite{graves2017automated}, curriculum through self-play~\cite{sukhbaatar2017intrinsic}, automatic goal generation~\cite{florensa2018automatic}, and skill-based curriculum~\cite{jabri2019unsupervised}.
We design a task-specific curriculum to train the RL agent for the rule generation task.

\section{Conclusion and Future Work}
In this paper, we propose a two-phased approach based on reinforcement learning to mine rules from KG. 
We first train an RL agent to generate quality rules rewarded by distributed representation, and then we use the trained agent to guide the rule search in the rule mining procedure.
We conduct extensive experiments to prove that our approach has state-of-the-art performance in both efficiency and effectiveness.

There are some future works towards rule mining via RL.
First, the rules we mine are limited in CP rules. It is worthy to explore the possibility of mining different kinds of Horn-clause rules with RL.
Second, we only use the embedding measures to train our RL agent. How to effectively and efficiently combine the statistical measures and embedding measures in reinforcement learning is also a meaningful direction to explore.
We expect this initial work to serve as a basis of comparison and inspiration for the development of rule mining via reinforcement learning.

\bibliography{rl_rules_clean}

\end{document}